\documentclass[conference]{IEEEtran}
\IEEEoverridecommandlockouts
\usepackage{cite}
\usepackage{amsmath,amssymb,amsfonts}
\usepackage{algorithmicx}
\usepackage{graphicx}
\usepackage{textcomp}
\usepackage{xcolor}
\def\BibTeX{{\rm B\kern-.05em{\sc i\kern-.025em b}\kern-.08em
    T\kern-.1667em\lower.7ex\hbox{E}\kern-.125emX}}
    
\usepackage{amsthm}
\usepackage{subfig}
\usepackage{epstopdf}
\usepackage{algpseudocode}
\usepackage{algorithm}
\usepackage{setspace}
\usepackage[T1]{fontenc}
\usepackage{color,soul}
\usepackage{multirow}
\usepackage{array}
\usepackage{cite}

\usepackage{verbatim}
\usepackage{enumitem}
\usepackage{url}
\usepackage{csvsimple}
\usepackage{float}

\usepackage{float}
\usepackage{caption}
\usepackage[capitalize]{cleveref}
\usepackage{environ}

\usepackage{array}
\newcommand{\PreserveBackslash}[1]{\let\temp=\\#1\let\\=\temp}
\newcolumntype{C}[1]{>{\PreserveBackslash\centering}p{#1}}
\newcolumntype{R}[1]{>{\PreserveBackslash\raggedleft}p{#1}}
\newcolumntype{L}[1]{>{\PreserveBackslash\raggedright}p{#1}}

\begin{document}

\title{FetFIDS: A Feature Embedding Attention based Federated Network Intrusion Detection Algorithm}

\author{\IEEEauthorblockN{Shreya Ghosh\IEEEauthorrefmark{1}, Abu Shafin Mohammad Mahdee Jameel\IEEEauthorrefmark{1}, Aly El Gamal\IEEEauthorrefmark{1} }
\IEEEauthorblockA{\IEEEauthorrefmark{1} School of Electrical and Computer Engineering, Purdue University, USA}
\IEEEauthorblockA{Email: {\{ghosh64, amahdeej, elgamala\}@purdue.edu}} }

\maketitle

\begin{abstract}
Intrusion Detection Systems (IDS) have an increasingly important role in preventing exploitation of network vulnerabilities by malicious actors. Recent deep learning based developments have resulted in significant improvements in the performance of IDS systems. In this paper, we present FetFIDS, where we explore the employment of feature embedding instead of positional embedding to improve intrusion detection performance of a transformer based deep learning system. Our model is developed with the aim of deployments in edge learning scenarios, where federated learning over multiple communication rounds can ensure both privacy and localized performance improvements. FetFIDS outperforms multiple state-of-the-art intrusion detection systems in a federated environment and demonstrates a high degree of suitability to federated learning. The code for this work can be found at https://github.com/ghosh64/fetfids.


\end{abstract}
\IEEEpeerreviewmaketitle
\section{Introduction}

The rise of Internet of Things (IoT) based systems has created more opportunities for web attackers to exploit vulnerabilities in a network and obtain confidential data. Intrusion Detection Systems (IDS) have become a critical component of networks, and have generated significant research interest. Traditional methods such as Naive-Bayes classifiers, Random Forest classifiers, and Support Vector Machines (SVM) have been long instrumental in developing robust IDS algorithms \cite{koc2012network}. However, multiple studies show data-driven learning based methods to outperform traditional methods \cite{alkasassbeh2016detecting, vinayakumar2019deep}. 

Recent development in the IDS domain center around deep learning algorithms \cite{barnard2022robust,autoenc,transactions}. In \cite{autoenc}, an autoencoder based intrusion detection system is proposed that can outperform popular SVM approaches by a large margin. The advent of attention based deep learning---which has been shown to perform very well in the computer vision and natural language processing domain---has created significant research interest and has resulted in the application of vision transformers in intrusion detection \cite{flow_transformer}. In \cite{robust_transformer,transformer_label,transformer_attention,transformer_lstm}, transformer based models have been shown to provide superior intrusion detection performance in diverse network environments. 

Another recent advancement in Intrusion Detection Systems is the adoption of federated learning. Federated approaches, especially for IoT devices, demonstrate superior performance compared to centralized machine learning-based intrusion detection systems \cite{rahman2020internet}. In \cite{transactions} a multinomial logistic regression (MLR) based method is shown to perform well in a federated learning environment. In \cite{ghosh2023an}, a transferable federated learning based IDS is presented.

\begin{figure*}[t]

    \centering
    \includegraphics[width=0.9\textwidth]{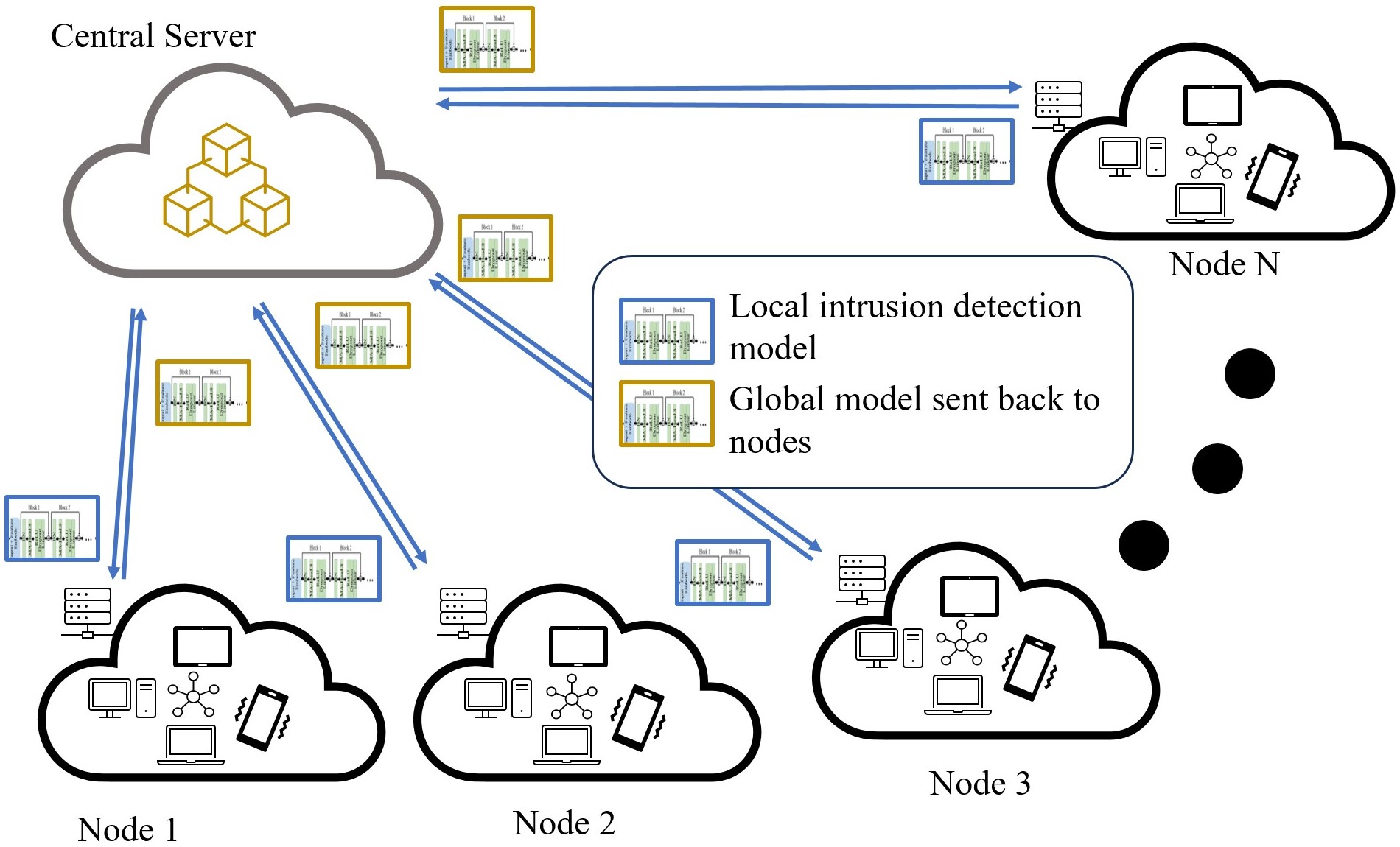}
    \caption{A summary of the federated setup used to train models deployed on the connected IoT devices. The nodes represent the devices chosen to participate in federated training and the local models are the models used on the devices to train with their local, private data. \vspace{-10pt} }
    \label{fig:system overview}  
\end{figure*}

\begin{figure}[t]
\includegraphics[width=0.4\textwidth]{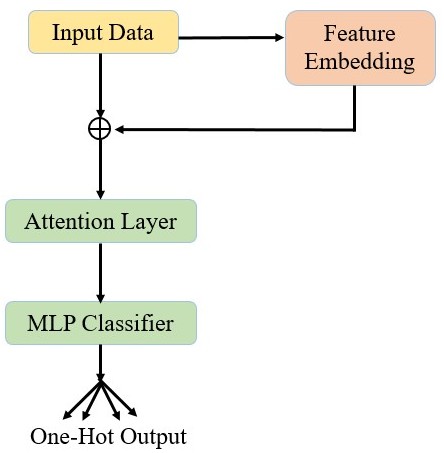}
\caption{Architecture of the proposed intrusion detection system. \vspace{-15pt}}
    
\label{fig:whole_architecture}
    
\end{figure}

In this paper, we develop Feature embedded transformer based federated IDS (FetFIDS), a novel attention based deep learning model utilizing feature embedding. we propose a federated learning regime with the explicit aim of enhancing intrusion detection for each deployed node and compare our model's performance to benchmark algorithms. 

The contributions of this work are as follows:
\begin{enumerate}
    \item We develop a novel transformer based deep learning model aimed at focusing the attention of the model on identifying unique features of attack data. To the best of our knowledge, this is the first use of feature embedding instead of positional embedding to improve intrusion detection performance. 
    \item We also propose the use of sequential attention blocks instead of a single attention block to further improve the performance. Our algorithm outperforms multiple state-of-the-art algorithms in a federated intrusion detection environment.
    \item Re-implementing deep learning based IDS benchmark methods can be challenging due to the absence of publicly available code for a vast majority of the published algorithms. We have made code implementations of both our proposed method, and the used benchmark methods publicly available. This ensures reproducibility of the presented results, and can be useful to future researchers in this domain. 
\end{enumerate}

This paper is organized in the following manner: In Section \ref{systemarch}, we explain our system architecture, the feature embedding process, the classifier model, and the federated intrusion detection environment. Finally, in Section \ref{results}, we present our experimental setup and show that our proposed algorithm significantly improves the intrusion detection performance in a federated environment.

\section{System Architecture}\label{systemarch}

In Fig.\ref{fig:system overview}, we present the setup for our federated intrusion detection system. Then in Fig. \ref{fig:whole_architecture}, we present the complete architecture of our proposed deep learning model, incorporating an attention based network with feature embedding, and an multi layer perceptron based classifier. In this section, we further discuss the different components of the setup.

\subsection{Nodes}

We consider a network of Internet of Things devices connected to each other. Each node monitors the data packets that are flowing through the network and detect malicious data packets injected by attackers using a deep learning based intrusion detection model deployed on the local devices. The intrusion detection systems are always collecting local data, training their models on local data and participating in communication rounds with the central server.  

\subsection{Attention Based Deep Learning Model}

\begin{figure*}[t]

\includegraphics[width=0.9\textwidth]{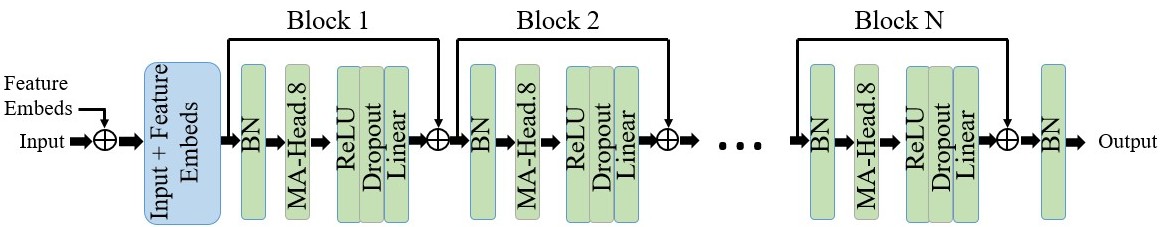}
\caption{Attention encoder. \vspace{-10pt} }
    
\label{fig:attention_encoder}
    
\end{figure*}

\begin{figure}[t]

\includegraphics[width=0.4\textwidth]{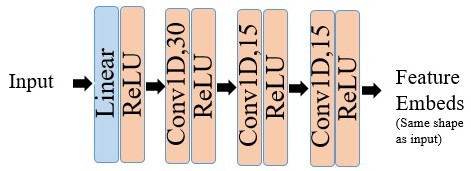}
\caption{Feature embedding architecture. \vspace{-10pt}}
    
\label{fig:feature_embedding}
    
\end{figure}

The deep learning model is an attention based model, presented in Fig. \ref{fig:attention_encoder}. The input is a 1D vector that contains the features of the network traffic data. Attention models are typically aimed at drawing the focus of the model on the contextually important parts of an input. The multihead attention function consists of several attention heads that focus the attention of the model on different aspects or different features of the data. The computed attention matrix tells the model which part of the input is important in relation to itself. This is called self-attention. Multiple attention heads (MA-Head) generate multiple attention matrices that enable the model to learn which aspects are important and which are not meaningful for model learning. The attention function itself takes three inputs: Query, Key. and Value. The query corresponds to what we are looking for, the key corresponds to what we can offer and the value corresponds to what we have. The input is broken down into these three vectors before it serves as inputs to the attention function.

\begin{equation}
    A(Q,K,V)=Softmax(\frac{QK^T}{\sqrt{d_k}})V
\end{equation}

where $d_k$ is the dimension of the Query and Key vectors. This process is repeated for each attention head. The softmax term is responsible for generating the actual attention matrix that tells the model which features most closely relate to which other features from the input. The matrix V then generates more contextually aware vectors that are the same shape as the input. Although the dimensions of the input and output are the same, the output vector is now a new value vector that contains more context about the inputs, related features and focuses on features that are important for model learning. We feed this vector into a learnable linear layer to consolidate the information and this is the output of the encoder model.

To ensure our model actually learns, we design the architecture of the model to have sequential multiheaded attention layers. This enables the contextually rich vector from each attention model to be further enriched by sequential computation of attention. This gives the model consolidated information about several aspects of the input. In addition, we use BatchNorm as our normalization method instead of LayerNorm which is usually used in attention based encoders. BatchNorm normalizes each feature independently depending on the batch statistics. This has been empirically shown to help our model perform with a higher accuracy of detection. The computation can be given as:

\begin{equation}
    \hat{x_i} \leftarrow \frac{x_i-\mu_B}{\sqrt{\sigma_B^2+\epsilon}}
\end{equation}
and
\begin{equation}
    y_i\leftarrow \gamma\hat{x_i}+\beta
\end{equation}
where $\mu_B \leftarrow \frac{1}{m}\sum_{i=1}^mx_i$ and $\sigma_B^2\leftarrow\frac{1}{m}\sum_{i=1}^m (x_i-\mu_B)^2$ and $m$ is the batch size. We use 8 attention heads and 3 attention blocks in our implementation.

\subsection{Feature Embedding} 
Attention models are typically aimed at drawing the focus of model learning towards a particular feature in a dataset. For a computer vision task, particular regions of an image could require model attention. Positional embeddings of the patches also tell the model about the location of those features. Similarly, for a language model, the position of a word in a sentence gives models information about positional context. However, for a network dataset, positional embeddings would not give the model very much contextual information because they typically have only one entry in the sequence, not a series of flows that are sequential and tell about the progression of the attack. Therefore, instead of using a positional embedding, we use what we term as a feature embedding. The deep neural network used for feature embedding is presented in Fig. \ref{fig:feature_embedding}. We use a Convolutional Neural Network (CNN) to generate the feature embedding. The model consists of sequential convolution layers and ReLU non linear activation. This enables the model to obtain high level features from the input. 
\begin{equation}
    y_j=b_j+\sum_{c=0}^{n_c-1}\sum_{k=-p}^{p}x_{c,j-k}w_{c,k}
\end{equation}

where $x$ is the input and $w$ are the weights corresponding to the kernel.
This feature embedding is combined with the original 1D input vector before they are both passed into the attention-based encoder layer.

\subsection{MLP Classifier} 

The MLP Classifier is a shallow classifier in which each layer performs the following computation:
\begin{equation}
    FFN(x)=max(0,xW+b)
\end{equation}
where $W,b$ are the weights and biases of that layer. This is followed by a Batch Normalization layer to provide stability to model training. The final output of the layer is converted into a probability distribution using the softmax activation function. This can be given as:
\begin{equation}
    Softmax{(z_i)}=\frac{e^{z_i}}{\sum_j^J e^(z_j)}
\end{equation}
where $J$ is the number of classes and $z_i$ corresponds to the layer output without activation. The output from this layer is used to compute the loss of the model. For our multiclass classification problem, we use Negative Log Likelihood loss and that can be given as:
\begin{equation}
    \theta_{ML}=\underset{\theta}{argmin} -E_{X,Y\sim P_{data}}log(P_{model} (Y|X,\theta))
\end{equation}

where $P_{data}$ is the distribution of the data, $P_{model}$ accounts for uncertainities in the model.

\subsection{Federated Setup for Intrusion Detection}
Each node trains a local model using only locally available data. Only the local model weights are then sent for aggregation to the central server. The central server employs a FedAvg aggregation algorithm \cite{fedavg} to generate parameter weights for a global model. This averaged global model is then sent back to all the nodes. The nodes then train this global model further using local training data. This process is repeated for multiple communication rounds.

The operation to calculate global model weights from local model weights can be represented by the following equation:

\begin{equation}
    \forall k, w_{t+1}^k \leftarrow w_t - \eta g_k; w_{t+1} \leftarrow \sum_{k=1}^{K} \frac{n_k}{n} w_{t+1}^k 
\end{equation}
where $w_t$ are the model weights after communication round $t$, $n_k$ is the number of local samples in the training data, assuming we have $K$ total nodes.

\section{Experimental Results}\label{results}

\subsection{Dataset}

In this study, we use the NSLKDD dataset \cite{nslkdd}, which is a popular standardized network traffic dataset, generated using traffic monitoring software to log and monitor network traffic packets. This dataset has 41 features and 5 different kinds of attack classes broadly classified into Benign, DoS, U2R, R2L, and Probing. We divide the training data among 5 devices with the same distribution of data in each device. This leads to class imbalance for some of the classes of data, however, we do not augment the data to address the imbalance. This is to emulate a more realistic scenario where we cannot guarantee the number of data packets seen at each device, and a model may not see many data packets belonging to a certain class. We use 85\% of the data for training, 7.5\% for validation, and 7.5\% for testing.

\subsection{Experimental Setup}

The main aim of this paper is to improve the accuracy of detection of malicious data packets. During training, each node sees all classes of data for classification. There is also no overlap between the benign or attack data present in different nodes during training. Each node only sees a fraction of the training data, effectively making the the number of samples in the local datasets much lower than a centralized system. We employ the AdamW \cite{adamw} optimizer with a learning rate of 0.001, weight decay of 1e-2 and the loss function is Categorical Crossentropy. We use a learning rate scheduler that follows an exponential decay. For the federated model, we run 20 rounds of federated model aggregation with 20 epochs of local training. We test the performance of intrusion detection algorithms after each communication round using a separate test set that has no overlap with the training or validation set. 

\subsubsection*{Benchmark Algorithms}

In this study, we inspect and compare intrusion detection performance for the proposed method with multiple state-of-the-art algorithms. Firstly, we implement the autoencoder based method presented in \cite{autoenc}. In this approach, an autoencoder is trained to learn the important features of the input data. The important features are then extracted from the autoencoder and fed into a MLP classification head to perform multiclass classification. This method was shown to outperform traditional SVM and Robust Covariance based classifiers by a significant margin in a centralized learning environment. The second benchmark implementation is TransIDS, a transformer based model similar to \cite{transformer_label} which has been shown to perform well in a centralized IoT environment. Finally, we also compare against a multinomial logistic regression (MLR) based method from \cite{transactions} that used the SGD classifier, and was shown to achieve state-of-the-art performance in a federated learning environment.

\begin{figure}[t]
   \centering
   \includegraphics[width=0.48\textwidth]{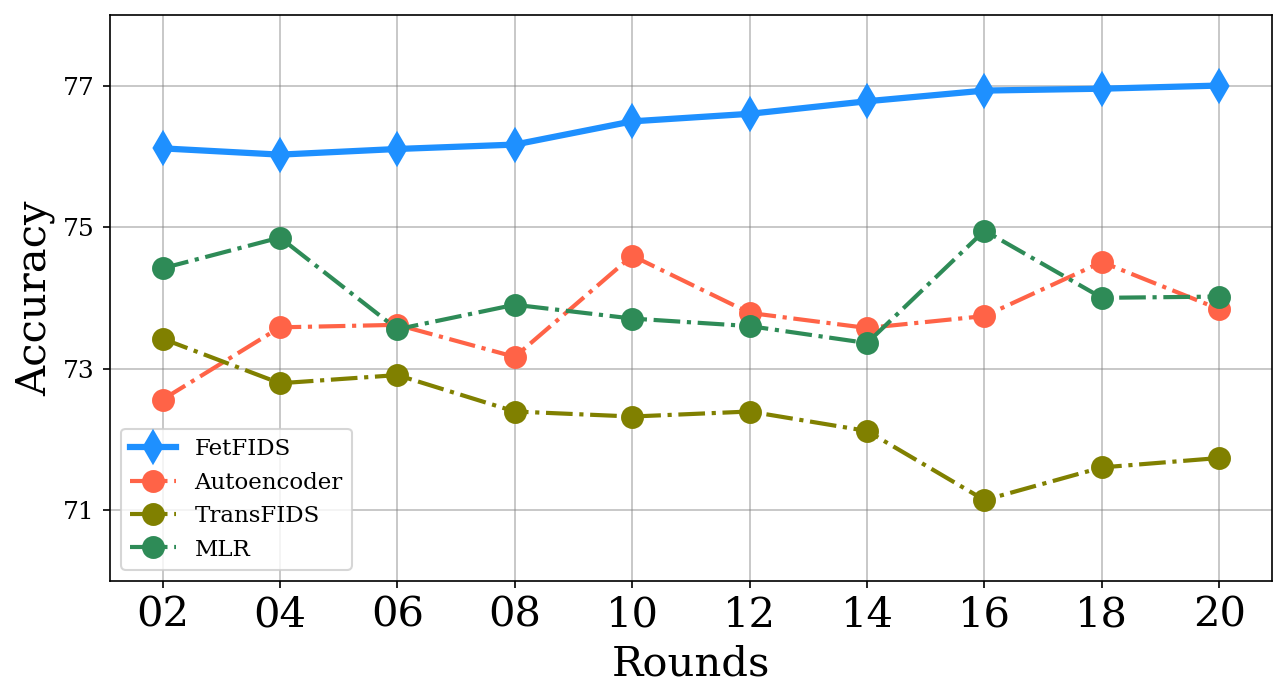}
   \caption{Comparison of performance between the Proposed method, Autoencoder, TransFIDS, and MLR over 20 federated communication rounds.}
   \label{fig:accuracy}
\end{figure}

\noindent
\begin{table}[t]
    \centering
    \captionsetup{justification=centering}
    \caption{Performance Characteristics for Different Models.}
    \label{table:performance}
    \begin{tabular}{l|c|c|c|c}
    \hline
        & Accuracy& Precision& Recall& F1 Score\\ 
        \hline
        FetFIDS& \textbf{77.00} & \textbf{97.47} & \textbf{64.66} & \textbf{77.73}  \\ 
        \hline
        Autoencoder& 73.84 & 96.68 & 60.06 & 74.06 \\ 
        \hline
        TransFIDS& 71.74 & 97.05 & 56.91 & 71.68 \\
        \hline
        MLR& 74.01& 96.36& 60.34& 71.17\\
        \hline
    \end{tabular}
\end{table}

\subsection{Performance Comparisons}

\begin{figure}[t]
   \centering
   \includegraphics[width=0.48\textwidth]{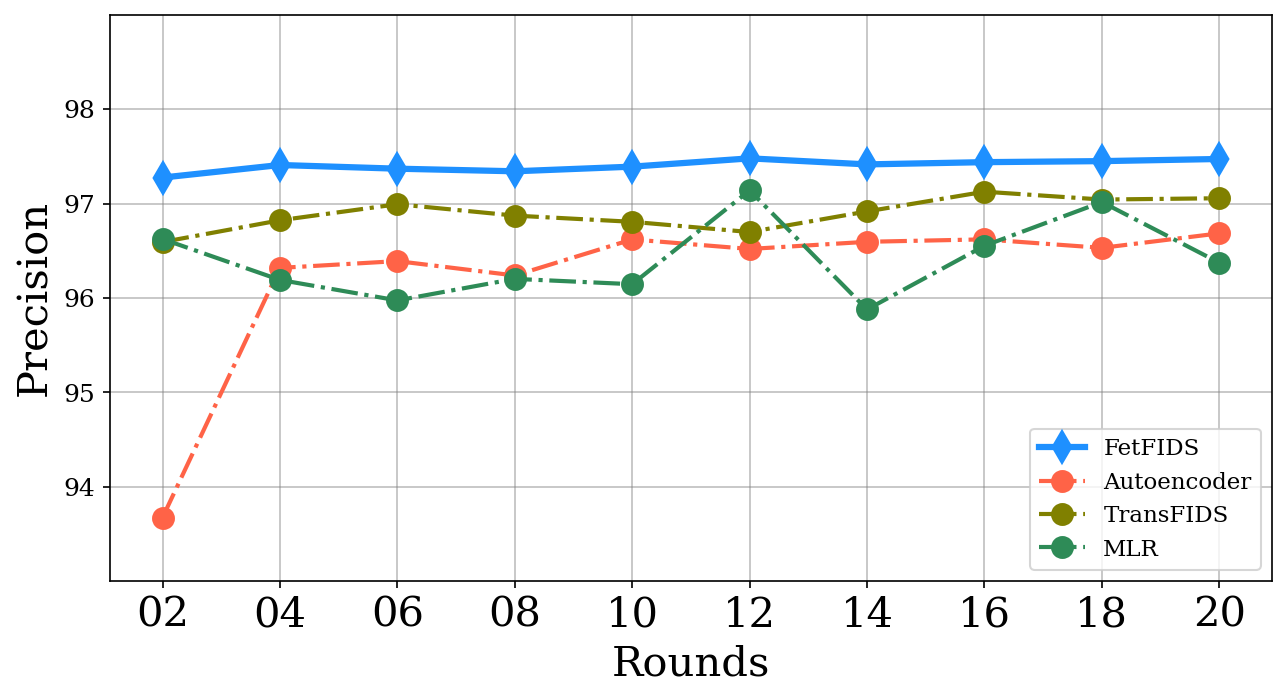}
   \caption{Comparison of precision values over 20 federated communication rounds. \vspace{-10pt}}
   \label{fig:precision}
\end{figure}

\begin{figure}[t]
   \centering
   \includegraphics[width=0.48\textwidth]{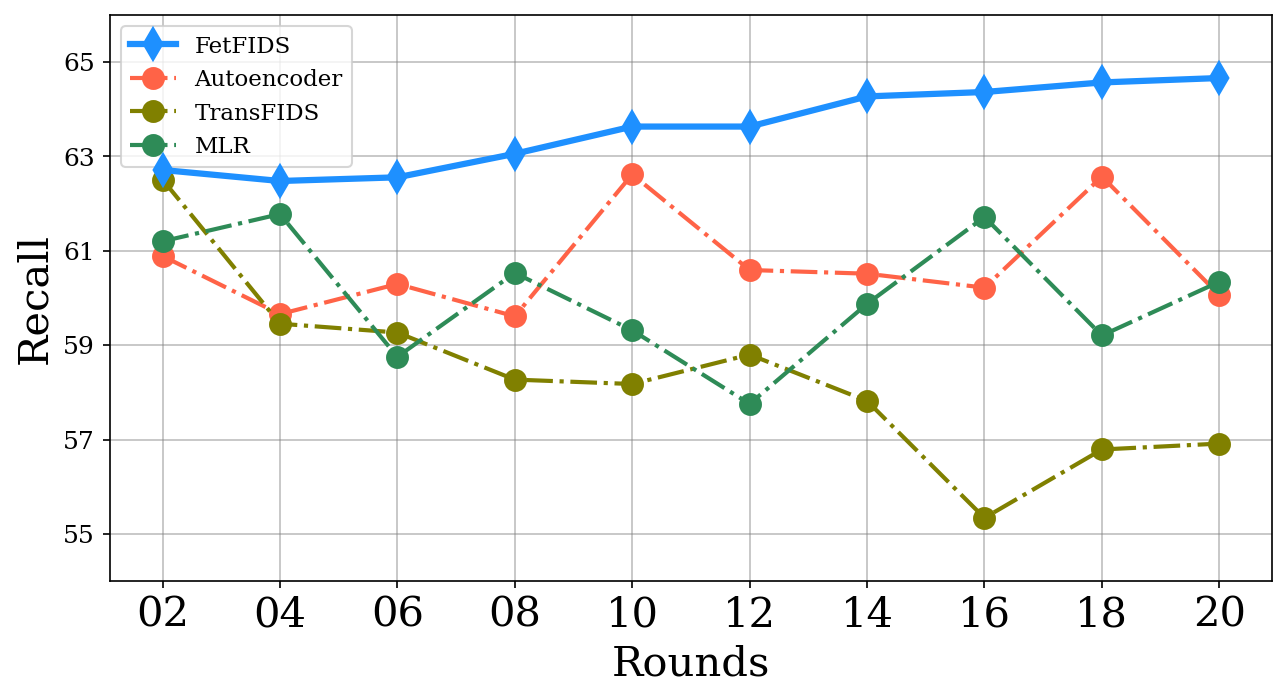}
   \caption{Comparison of recall values over 20 federated communication rounds. \vspace{-10pt}}
   \label{fig:recall}
\end{figure}

\begin{figure}[t]
   \centering
   \includegraphics[width=0.48\textwidth]{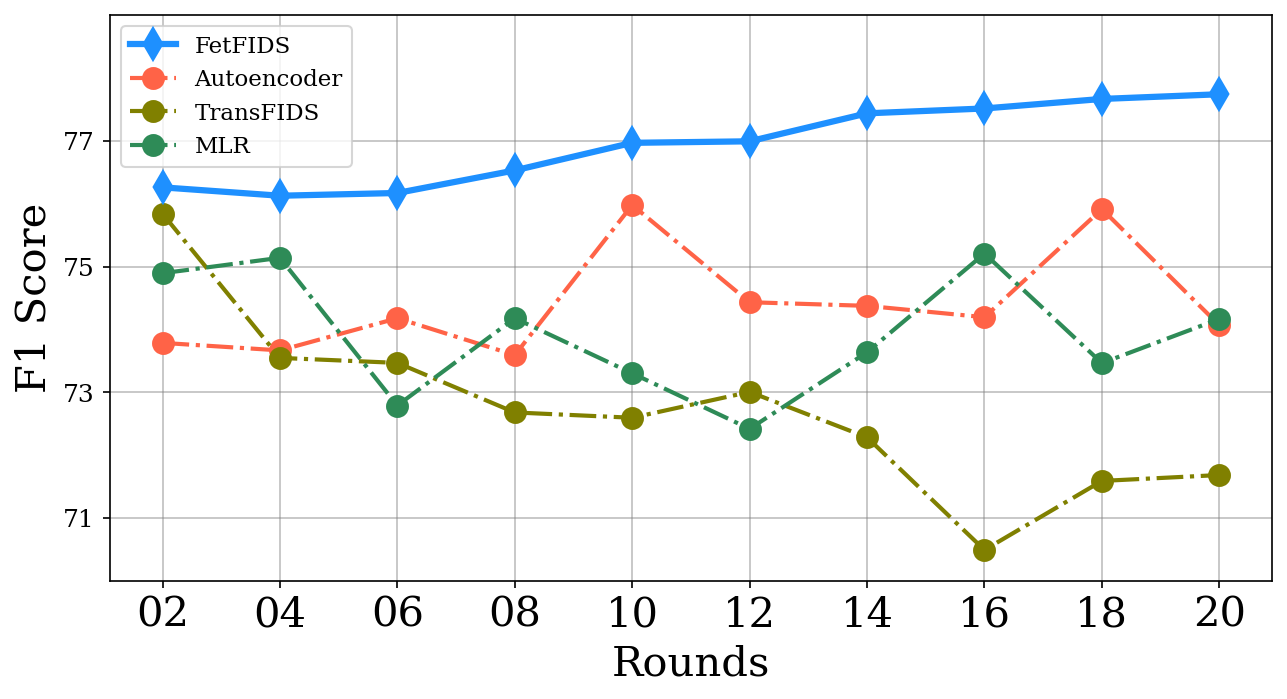}
   \caption{Comparison of F1 values over 20 federated communication rounds. \vspace{-10pt}}
   \label{fig:f1}
\end{figure}

In Fig. \ref{fig:accuracy} we present the accuracy of FetFIDS and the three benchmark methods over the 20 training rounds. In Table \ref{table:performance}, we present the final values of the four performance metrics after 20 communication rounds. Then in Figs. \ref{fig:precision}, \ref{fig:recall}, and \ref{fig:f1} we track the precision, recall and F1 score for the same setups. 

From Figs. \ref{fig:accuracy}, \ref{fig:precision}, \ref{fig:recall}, and \ref{fig:f1}, we can see that FetFIDS outperforms the competing methods on all four metrics - accuracy, precision, recall, and F1 score. From Fig. \ref{fig:accuracy}, we see that in terms of intrusion detection accuracy, there is a visible difference between FetFIDS and the autoencoder, TransIDS, and MLR based methods. Among the compared methods, our method beats the best performing method (MLR) by 3\% and the lowest performing method (TransFIDS) by 7\%. FetFIDS shows an increase in accuracy with increase in communication rounds which is the desired behaviour of the model, indicating that the model learns new information over communication rounds. Our proposed model exhibits stable testing accuracies and less fluctuations over the communication rounds, beating the other methods in every communication round.

For TransIDS, the performance deteriorates with communication rounds, which is surprising as it is also an attention-based method.  Our implementation of multi headed attention blocks follows closely the architecture of the encoder in a transformer block, however, the difference is that we use a feature embedding module instead of a positional encoding model, and utilize sequential attention blocks. This goes to show that the vector being enriched with sequential feature-based information is more meaningful to the model than positional or contextual information. 

The performance of TransIDS highlights the risks associated with blindly employing models developed for a centralized learning environment to a federated environment. On the other hand, despite also being developed for a centralized system, the autoencoder performs almost at par with MLR, which was developed for a federated system.

From Fig. \ref{fig:precision}, we can see that FetFIDS is very good at avoiding false positives. Recall, which is the most challenging of the four metrics, shows good performance improvements for the proposed method over federated communication rounds in Fig. \ref{fig:recall}. The other algorithm specifically developed for a federated setup, MLR, shows instability in performance, where it fails to maintain performance improvements over communication rounds. Finally, from Fig. \ref{fig:f1}, the F1 score shows a similar trend to the accuracy plot, but the improvement in the performance of FetFIDS over multiple communication rounds is more evident here.

\begin{figure}[t]
   \centering
   \includegraphics[width=0.48\textwidth]{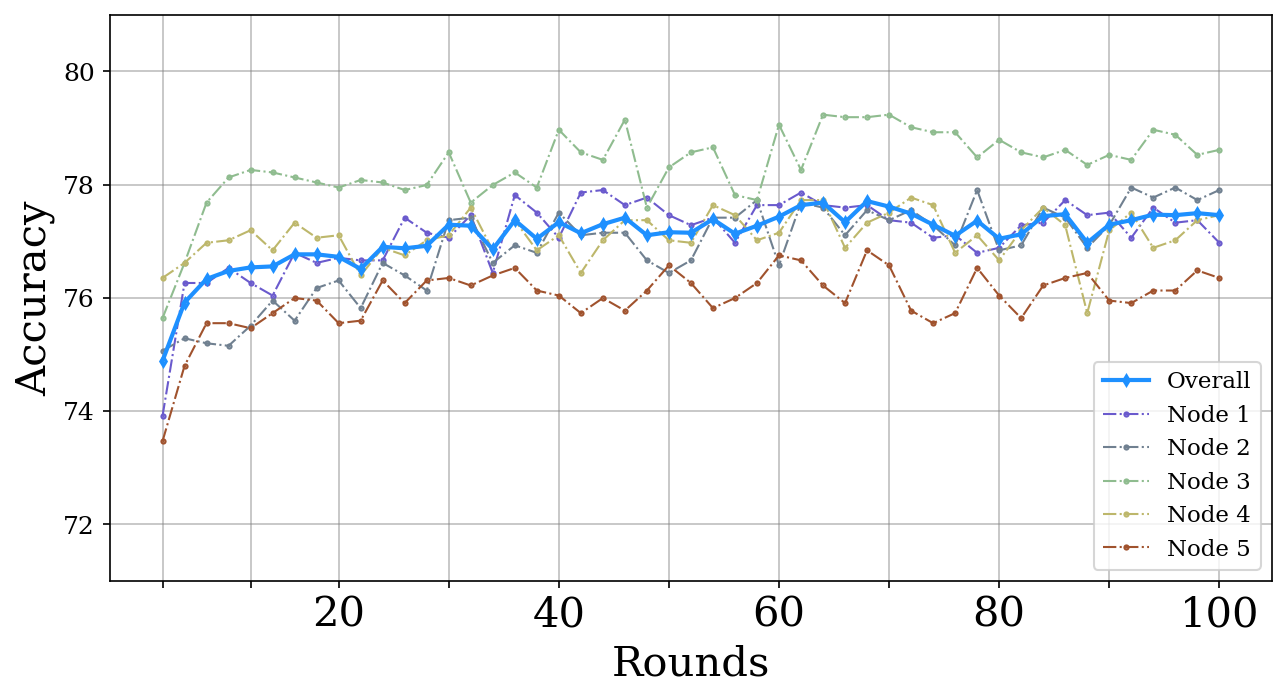}
   \caption{Intrusion detection accuracy for all 5 nodes over 100 federated communication rounds. \vspace{-10pt}}
   \label{fig:Proposed_100}
\end{figure}

\subsection{Node Performance}

Next, in Fig. \ref{fig:Proposed_100}, we present the per node accuracies for FetFIDS for a simulation run where we let the system train for 100 communication rounds, and provide the intrusion detection performance on the test set after each communication round. We can see that the proposed method provides stable learning improvements for all the nodes. There is some difference between the accuracies of different nodes, with Node 3 showing consistently better performance and Node 5 showing consistently worse performance compared to the average. Overall we can see that there are clear phases of rapid learning followed by somewhat stable results. This indicates that for resource constrained training setups, reducing the number of federated communication rounds can be a good trade-off between performance and computational requirements.

\subsection{Hyperparameter Tuning}

\begin{figure}[t]
   \centering
   \includegraphics[width=0.48\textwidth]{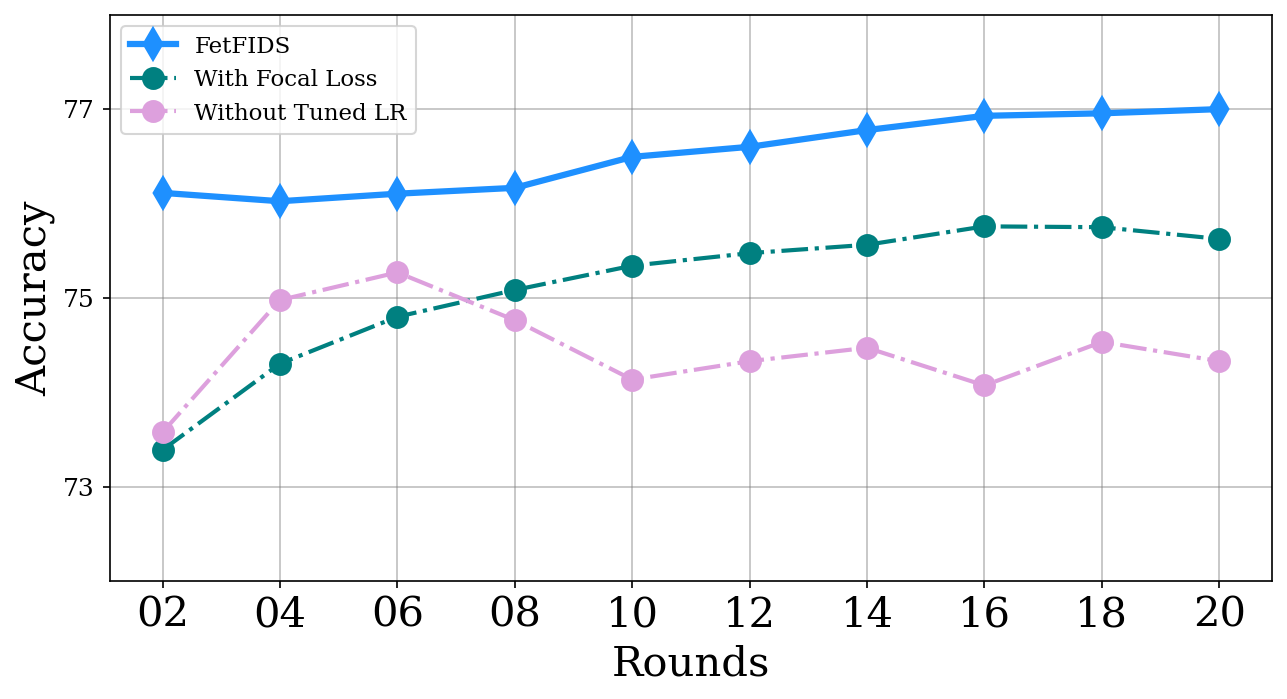}
   \caption{Effect of hyperparameter tuning.}
   \label{fig:hyperparameter}
\end{figure}

In Fig. \ref{fig:hyperparameter}, we present the empirical evidence behind some of the hyperparameter choices in this work. Firstly, we explore the effect of tuning the learning rate. For FetFIDS, we use exponential learning rate decay with gamma of 0.7 and categorical crossentropy loss, selected via empirical experiments. We present another case with a fixed learning rate of 0.0001 without exponential decay (labeled 'Without Tuned LR' in Fig. \ref{fig:hyperparameter}). This has worse performance and also demonstrates learning instability between communication rounds. Next, we investigate the use of focal loss\cite{focalloss} instead of categorical crossentropy, as this is supposed to be advantageous in a federated dataset with in-node class imbalance. However, focal loss fails to bring performance improvements.

\subsection{Computational Complexity}
In Table \ref{table:complexity}, we present the Floating Point Operations Per Second (Flops) and number of parameters for the three deep learning based models and the inference time for a single test packet for the four algorithms. Firstly, in terms of local processing power, the FetFIDS model needs more computing power compared to the Autoencoder or TransFIDS. This is due to the employment of multiple attention blocks, and the feature embedding pipeline. However, TransFIDS has the highest number of model parameters, which means that it requires the most bandwidth for federated model aggregation. Finally, in terms of inference time, MLR is the fastest one, which is expected as it is a simple logistic regression based method. All of the methods are very fast, processing a single packet in a matter of microseconds. This shows that the performance improvements provided by the feature embedding based approach have some trade-offs in terms of inference speed and computational complexity.

\noindent
\begin{table}[t]
    \centering
    \captionsetup{justification=centering}
    \caption{Number of Parameters, Flops, and Inference Time for Different Algorithms.}
    \label{table:complexity}
    \begin{tabular}{l|c|c|c|c}
    \hline
        & FetFIDS& Autoenc.& TransFIDS& MLR\\
        \hline
        Flops& 1.07M& 0.94M& 0.16M& -\\ 
        \hline
        Parameters& 116.64k& 66.51k& 121.70k& - \\ 
        \hline
        Time  &19.10$\mu s$ &4.67$\mu s$ & 14.90$\mu s$ & 0.24$\mu s$ \\
        \hline
    \end{tabular}
\end{table}

\section{Conclusion}
In this paper, we develop an intrusion detection system for a federated learning environment. The proposed sequential transformer attention block based deep learning model incorporates feature embedding to generate more meaningful inputs to the intrusion detection classifier. The proposed algorithm, FetFIDS, outperforms benchmark intrusion detection systems while also demonstrating learning stability over federated learning communication rounds. Our model as well as our implementations of benchmark algorithms are publicly available for reproduction. In the future, we aim to focus on further developing the model to reduce its computational footprint and inference time, while maintaining state-of-the-art intrusion detection performance.

\ifCLASSOPTIONcaptionsoff
  \newpage
\fi
\bibliographystyle{IEEEtran} 

\bibliography{nattack2023}

\end{document}